# Pixel Super-Resolved Fluorescence Lifetime Imaging Using Deep Learning

Paloma Casteleiro Costa[1,2], Parnian Ghapandar Kashani[1], Xuhui Liu[3], Alexander Chen[1], Ary Portes[1], Julien Bec[3], Laura Marcu[3], and Aydogan Ozcan[1,2,4,*]

[1]Electrical and Computer Engineering Department, University of California, Los Angeles, CA, 90095, USA
[2]Bioengineering Department, University of California, Los Angeles, CA, 90095, USA
[3]Department of Biomedical Engineering, University of California, Davis, California 95616, USA
[4]California NanoSystems Institute (CNSI), University of California, Los Angeles, CA, 90095, USA
[*]Correspondence to: ozcan@ucla.edu

**Abstract**

Fluorescence lifetime imaging microscopy (FLIM) is a powerful quantitative technique that provides metabolic and molecular contrast, offering strong translational potential for label-free, real-time diagnostics. However, its clinical adoption remains limited by long pixel dwell times and low signal-to-noise ratio (SNR), which impose a stricter resolution-speed trade-off than conventional optical imaging approaches. Here, we introduce $FLIM_{PSR\_k}$, a deep learning-based multi-channel pixel super-resolution (PSR) framework that reconstructs high-resolution FLIM images from data acquired with up to a 5-fold increased pixel size. The model is trained using the conditional generative adversarial network (cGAN) framework, which, compared to diffusion model-based alternatives, delivers a more robust PSR reconstruction with substantially shorter inference times, a crucial advantage for practical deployment. $FLIM_{PSR\_k}$ not only enables faster image acquisition but can also alleviate SNR limitations in autofluorescence-based FLIM. Blind testing on held-out patient-derived tumor tissue samples demonstrates that $FLIM_{PSR\_k}$ reliably achieves a super-resolution factor of $k = 5$, resulting in a 25-fold increase in the space-bandwidth product of the output images and revealing fine architectural features lost in lower-resolution inputs, with statistically significant improvements across various image quality metrics. By increasing FLIM's effective spatial resolution, $FLIM_{PSR\_k}$ advances lifetime imaging toward faster, higher-resolution, and hardware-flexible implementations compatible with low-numerical-aperture and miniaturized platforms, better positioning FLIM for translational applications.

# 1 INTRODUCTION

Fluorescence lifetime imaging microscopy (FLIM) is a powerful quantitative technique for mapping the excited-state decay kinetics of endogenous and exogenous fluorophores in biological tissue[1-3]. Because fluorescence lifetime is largely independent of fluorophore concentration and illumination intensity, FLIM offers robust biochemical contrast and supports applications such as metabolic imaging, redox analysis, and the study of molecular interactions[4-7]. High-resolution FLIM, however, remains inherently slow due to the long pixel dwell times required, the limited photon budget of endogenous autofluorescence, and the trade-offs among field of view, signal-to-noise ratio (SNR), and spatial sampling density[8,9]. These constraints are particularly limiting in thick, heterogeneous tissues such as those of the head and neck, where rapid mapping of metabolic gradients and subcellular morphology would benefit both basic research and emerging *in vivo* optical imaging technologies[10-13]. Although point-scanning systems can yield excellent spatial and lifetime accuracy, they are often too slow and impractical for many clinical and high-throughput imaging applications. In addition to scan time, achieving higher spatial resolution in FLIM can require longer tissue exposure, higher excitation energies, or more complex optical configurations to maintain adequate photon counts and lifetime precision[14,15]. These constraints are amplified in *in vivo* settings, where tissue motion, scattering, safety limits, and probe miniaturization impose strict limits on dwell times and collection efficiency.

Pixel super-resolution (PSR) offers an attractive solution to the speed-resolution bottleneck in FLIM. The term pixel super-resolution, as widely used in computational imaging and computer vision prior to the introduction of optical nanoscopy, denotes the reconstruction of images with a higher spatial sampling density and an increased effective space-bandwidth product, *within* the diffraction limit of the imaging system[16-24]. In this context, PSR refers to reconstructing higher-resolution images from undersampled or binned measurements, thereby increasing the number of useful pixels and enhancing structural interpretability without altering the underlying optics[25-27]. This computational approach has shown broad utility across biomedical imaging, including label-free tissue microscopy, where learning-based PSR methods have enabled several-fold gains in effective resolution by mapping coarse inputs to higher-resolution structural domains, as well as imaging mass spectrometry, where generative models have recovered histological detail from data acquired with substantially larger pixel sizes[28,29]. These studies illustrate how PSR can expand the practical resolution of various imaging modalities that are inherently limited by acquisition speed, photon budget, or instrument design constraints, which closely parallel those encountered in FLIM. Deep learning has become a central approach for PSR across different biomedical imaging modalities, particularly where physical acquisition of densely sampled measurements is impractical. Among these methods, conditional generative adversarial networks (cGANs)[30,31] have been widely used not only for super-resolution, but also for image translation and restoration tasks, owing to their ability to learn complex cross-domain mappings while preserving fine structural detail[25,32-34]. In FLIM, deep neural networks have enabled denoising, phasor-space regression, lifetime unmixing, and rapid parameter extraction[35-39].

Here, we introduce a deep-learning-based pixel super-resolution framework for FLIM that reconstructs high-resolution lifetime and intensity images from low-resolution FLIM inputs (**Figure 1**). Each input field of view consists of six channels: three lifetime channels and three autofluorescence intensity channels, captured from label-free tissue sections using a high-resolution point-scanning FLIM system. These high-resolution maps serve as ground truth, while the low-resolution images emulate faster, coarser FLIM acquisition with a larger scanning pixel size and reduced dwell times, mirroring practical constraints in clinical or *in vivo* scenarios. We employ a cGAN framework to learn the mapping between low-resolution and high-resolution FLIM data. While some recent pixel super-resolution works have demonstrated outstanding results with more advanced generative models, e.g., probabilistic diffusion models, cGANs offer several advantages for image super-resolution tasks, particularly in the biomedical context. First, they provide strong structural fidelity and edge preservation, where adversarial training helps the generator recover high-frequency content that would otherwise be lost[28-30,32]. Second, unlike diffusion models, which require hundreds of iterative denoising steps for each image inference and thus exhibit substantially longer computational latency, cGANs enable nearly instantaneous reconstruction after a single forward pass through the generator. This speed advantage is essential for any future translation of FLIM PSR into real-time microscopic or *in vivo* imaging workflows.

In this study, we systematically evaluate pixel super-resolution factors ($k$) from $k = 2$ to $k = 7$, with particular emphasis on the upper limit of stable PSR performance. Our results demonstrate that the cGAN-based models, denoted FLIM$_{PSR\_k}$, provide accurate structural and biochemical reconstructions up to $k = 5$, enabling a substantial improvement of the effective FLIM resolution without compromising lifetime accuracy. Beyond this regime ($k \geq 6$), both cGAN and diffusion-based PSR approaches become increasingly challenged, exhibiting characteristic artifacts and reduced lifetime precision. These findings are validated using whole-slide FLIM data from patient-derived head and neck tumor specimens, with strict patient-wise separation between the training and blind testing sets.

By substantially increasing the effective spatial resolution of FLIM without altering image acquisition hardware, this PSR framework has the potential to reduce FLIM scanning times by more than an order of magnitude, enabling rapid mapping of metabolic and microstructural features in human tissues. In head and neck pathology, where fine stromal organization, cellular morphology, and metabolic gradients are clinically informative, FLIM$_{PSR\_k}$ may facilitate faster *ex vivo* assessments and lay the groundwork toward future high-resolution *in vivo* FLIM imaging systems, complementing emerging fiber-optic and catheter-based lifetime technologies[11-13]. More broadly, this work extends the paradigm of deep-learning-based PSR to lifetime-resolved imaging and establishes a foundation for accelerated FLIM that maintains accuracy across multi-parametric channels while enabling substantial gains in resolution, throughput, and translational potential.

## 2 RESULTS

### cGAN-based FLIM pixel super-resolution

We blindly tested the performance of our cGAN-based FLIM pixel super-resolution framework on whole slide images of human head and neck tumor tissue samples from three held-out patients, yielding 128 test image patches with a 1.92 × 1.92 mm field-of-view each. We denote each model with FLIM$_{PSR\_k}$, where $k$ represents the pixel super-resolution factor and $k \in \{2,3,4,5,6,7\}$. For each PSR factor, the input data was obtained by block-averaging the high-resolution data with $k \times k$ blocks, resulting in a $k^2$-fold reduction in the number of input pixels. We employed a cGAN framework to learn the mapping between the low-resolution and high-resolution FLIM data; see the Methods section for details. The inference time for each 1.92 × 1.92 mm, six-channel FLIM image patch was ~107 ms (see the Methods section for details).

**Figures 2** and **3** present an evaluation of the performance of the FLIM$_{PSR\_k}$ framework at a pixel super-resolution factor of 5 (i.e., FLIM$_{PSR\_5}$) in both lifetime and intensity channels, respectively. Each figure shows multiple representative regions that illustrate the model's behavior across different tissue structures. For each region, we display the low resolution (LR$_5$) FLIM input, the corresponding interpolated image, the FLIM$_{PSR\_5}$ output, and the high resolution (HR) target (the ground truth). To better quantify the PSR performance of FLIM$_{PSR\_5}$, we provide overlaid cross-sections through various structures, compared with the interpolated low-resolution image and the ground truth. The super-resolved profiles closely match the ground truth, capturing sharp intensity transitions and preserving peak amplitudes that are smoothed or lost in the interpolated images. Notably, the dips and peaks in the profiles align well between the super-resolved and ground-truth images, demonstrating faithful reconstruction of fine structural details. The FLIM$_{PSR\_5}$ output demonstrates sharper features and preserves structural details that are lost through interpolation. More examples of the FLIM$_{PSR\_5}$ output results are shown in **Figures S1** and **S2**.

To further evaluate the model's performance in the spatial frequency domain, we also compared the radially averaged power spectra of the low-resolution images, the FLIM$_{PSR\_5}$ output images, and the high-resolution targets. This analysis was performed for both lifetime (**Fig. 2K, L**) and intensity channels (**Fig. 3K, L**). In the lifetime channel, the low-resolution FLIM images exhibit substantial attenuation of mid- and high-frequency content. In contrast, the FLIM$_{PSR\_5}$ output recovers the frequency components that closely match the ground truth across a broad range of spatial frequencies. The improvement is consistent across all three lifetime bands. We observe similar trends in the corresponding intensity channels (**Fig. 3**). The FLIM$_{PSR\_5}$ reconstructed image demonstrates a clear recovery of mid-frequency data that was lost in the low-resolution counterpart (**Fig. 3L**). These findings suggest that the FLIM$_{PSR\_5}$ model effectively restores the spatial information necessary for accurate biochemical interpretation.

In **Figures 4** and **5**, we further study the method's performance for PSR factors from $k = 2$ to $k = 7$, for both lifetime and intensity channels, respectively. Our analysis indicates that the FLIM$_{PSR\_k}$ framework can successfully reconstruct FLIM data up to a PSR factor of $k = 5$. However, at PSR factors beyond $k = 5$, the cGAN framework can no longer reliably restore high-resolution features and starts to generate some spatial artifacts. Performance degradation becomes evident at these higher PSR factors of $k > 5$, where fine structures become lost or indistinguishable due to resolution limitations. We also report quantitative metrics, which support the same conclusion. The learned perceptual image patch similarity (LPIPS)[40] metric and the structural similarity index metric (SSIM)[41] exhibit an apparent decline in performance at $k = 6$. This pattern is consistent with the missing features and degradations visible in the zoomed-in regions and cross-sections, reflecting the reduced information available in the heavily downsampled inputs at higher PSR factors. Additional examples of the FLIM$_{PSR\_k}$ performance across different PSR factors are presented in the Supplementary Information (**Figures S3** and **S4**).

These image comparisons, featuring several zoomed-in regions, cross-sections, and quantitative image quality metrics, demonstrate that the FLIM$_{PSR\_k}$ framework yields reliable super-resolution reconstructions up to a PSR factor of $k = 5$. Beyond that point, the available information in the low-resolution FLIM input becomes insufficient to support accurate recovery of spatial and biochemical structures, resulting in some spatial artifacts.

**PSR performance comparison between cGAN and diffusion models**

To evaluate the performance of our FLIM$_{PSR\_k}$ framework relative to a diffusion-based generative approach, we trained both models on the same dataset, with matching conditions. The diffusion-based PSR models followed a Brownian bridge formulation[42], and both models (diffusion and cGAN) were trained to map the same low-resolution FLIM inputs ($I^{input}$) to their corresponding high-resolution targets ($I^{target}$). **Figure 6** summarizes the comparison of the two generative approaches at a PSR factor of $k = 5$. For $k = 5$, the cGAN-based image reconstructions remain stable across various sample fields of view and across all six FLIM channels. The PSR output images of FLIM$_{PSR\_5}$ preserve the morphology and local contrast of various structures visible in the high-resolution ground-truth images and avoid introducing non-physical details. The diffusion-based reconstructions also improve on the low-resolution input images, but show a clear tendency toward over-sharpening in some regions and loss of structural coherence in others. These effects appear as small artificial edges or local contrast modulations that are not present in the target, ground-truth images. These differences are reflected in the line profiles shown in **Figure 6**, where the diffusion model occasionally produces peaks or transitions that are either too sharp (**Fig. 6H**), inaccurate (**Fig. 6F**), or slightly displaced relative to the ground truth (**Fig. 6B, D**). Quantitative evaluations across all metrics support these observations. At $k = 5$, the cGAN achieves statistically significantly higher SSIM and peak SNR (PSNR) and lower mean square error (MSE) and LPIPS values in the output FLIM images compared to the diffusion model.

At higher PSR factors of $k > 5$, both models deteriorate, but in different ways. The cGAN tends to miss fine structures as the input becomes too coarse to constrain the mapping, leading to incomplete or smoothed features. The diffusion model, on the other hand, increasingly hallucinates structures that do not correspond to any content in the high-resolution target FLIM images. This behavior produces sharper but less reliable images. As a result, at $k = 6$ and $k = 7$, the performance differences between the two generative models vary from region to region (**Fig. 7**). In some areas, diffusion achieves better pixel-level agreement by inserting high-frequency content, while in other regions the cGAN performs better by avoiding non-physical features. Quantitatively, there is no clear winner in the higher PSR regime of $k > 5$ (**Fig. S5**). Neither model produces FLIM reconstructions that can be considered robust at these higher PSR factors, which aligns with the degradation observed in the examples shown in **Figure 7**.

These findings suggest that the cGAN framework is generally a more reliable approach for FLIM pixel super-resolution. At $k \leq 5$, it delivers consistent PSR reconstructions with better alignment with the ground-truth FLIM data. At higher PSR factors, the available information in the low-resolution input FLIM images becomes insufficient for either generative method to accurately recover spatial or biochemical structures.

It is also important to note that diffusion-based inference incurs a substantially higher computational cost than the cGAN approach. On an NVIDIA GeForce RTX 3090 Ti, the cGAN inference required ~0.1 s per 1.92 × 1.92 mm, six-channel FLIM image patch, while the diffusion-based inference required ~78 s per patch.

## 3 DISCUSSION

The results of this study demonstrate that cGAN-based pixel super-resolution can significantly expand the range of conditions under which FLIM data can be acquired and interpreted, providing a computational route to resolutions that would otherwise require slower scanning, higher illumination power, and more restrictive optics. By recovering fine spatial and biochemical structures from spatially undersampled measurements, FLIM$_{PSR\_k}$ changes the balance between hardware and computation in FLIM, allowing imaging performance to be determined less by optical hardware constraints and more by the information content of the acquired signal. This shift is particularly relevant for settings where high-resolution lifetime imaging has traditionally been impractical, including fast scanning over large tissue areas, imaging through low-numerical-aperture (NA) or miniaturized optics, and future *in vivo* implementations where motion, scattering, and low photon counts limit dwell time and sampling density. Our demonstrated ability to reconstruct accurate lifetime contrast from coarser measurements suggests that PSR can serve as a bridge between the biochemical specificity of FLIM and the speed and form-factor constraints of clinical instrumentation, including fiber-based and endoscopic systems[8,11,13].

In this context, the choice of a deep learning model is critical. Biomedical imaging differs from natural-image super-resolution not only in the types of contrast involved, but in the consequences of reconstruction errors. Minor hallucinations or misplaced boundaries may be tolerable in photography-related applications, but they are unacceptable in settings where images inform diagnostic or surgical decisions. Diffusion models have shown strong performance across diverse imaging tasks[26,43-45]. Yet, their iterative sampling process can introduce fine-scale variations that do not correspond to the underlying tissue structure unless carefully controlled and regulated[46]. In addition, and highly relevant for FLIM, diffusion inference is a slow process. Each output requires hundreds of sampling steps, resulting in image inference times that are typically much longer than those of a single forward pass through the generator of a cGAN-based approach. This difference is not simply a matter of computational efficiency; it determines whether PSR can be integrated into real-time or near-real-time FLIM workflows, including video-rate FLIM or *in vivo* lifetime imaging, where processing delays translate directly into reduced clinical usability.

In contrast, the cGAN-based PSR approach presented here provides stable and reliable reconstructions within a PSR factor of $k \leq 5$. Up to a PSR factor of $k = 5$, the $FLIM_{PSR\_k}$ model recovers structural details and lifetime gradients in close agreement with high-resolution reference images, digitally enabling a 25-fold gain in effective spatial sampling. The adversarial framework yields consistent results across FLIM channels, with predictable failure modes when the input no longer contains sufficient information (e.g., for $k > 5$) to support accurate image reconstruction. This behavior is preferable in a biomedical context, where conservative degradation is less risky than introducing non-physical features that look sharp, but in reality represent hallucinations[46]. The combination of structural fidelity and near-instantaneous inference positions cGAN-based $FLIM_{PSR\_k}$ as a powerful choice for FLIM PSR image reconstructions, particularly in settings where FLIM images may directly inform clinical decision-making.

Rather than a single $FLIM_{PSR\_k}$ model, one could also consider training a separate PSR model for each spectral FLIM channel. However, joint training across FLIM channels provides more stable and consistent image reconstructions than per-channel training; see **Fig. S6**. Independent networks may introduce uncorrelated errors that randomly create channel-to-channel inconsistencies, whereas a joint FLIM PSR model better leverages the shared spatial and structural information of different FLIM channels to reconstruct high-frequency details during the super-resolution image inference task. In our experiments, the single-channel models did not outperform the multi-channel approach, as illustrated in **Fig. S6**.

Future extensions of this work may incorporate physics-informed regularization constraints and/or uncertainty estimation approaches to further improve our FLIM PSR image reconstruction performance. Increasing dataset diversity, particularly across different tissue types and lifetime distributions, may also enhance the generalizability of the $FLIM_{PSR\_k}$ framework. Nonetheless, the presented results demonstrate that the adversarially trained $FLIM_{PSR\_k}$ framework can significantly improve the practical resolution of FLIM by a factor of five without compromising biochemical accuracy, offering a foundation for 25x faster, more versatile lifetime imaging in both research and translational settings.

Taken together, these results demonstrate that pixel super-resolution is a promising strategy for accelerating FLIM and enabling higher-resolution biochemical imaging in settings where hardware-based improvements are impractical. By offloading part of the resolution burden from optics to computation, $FLIM_{PSR\_k}$ may help reduce scan times, increase throughput, and broaden the applicability of FLIM to more challenging environments. At the same time, the comparison between cGAN and diffusion model-based generative approaches illustrates the importance of controlling hallucinations[46] and model variance in biomedical image super-resolution tasks. Approaches that maintain a conservative, structurally faithful behavior, particularly when the input information becomes highly limited, are better suited for clinical translation, where interpretability and reliability are essential. The presented $FLIM_{PSR\_k}$ approach has the potential to broaden the applicability of FLIM across research and clinical settings. Faster scanning, reduced SNR constraints, and compatibility with lower-NA or miniaturized optics could make FLIM PSR feasible in contexts where hardware-based improvements are impractical.

## 4 METHODS

**Imaging setup and FLIM data acquisition**

The FLIM data used in this study were acquired using a laser-scanning pulse-sampling FLIM (LSPS-FLIM) system[5]. The instrument (**Fig. 8**) employs a pulsed UV excitation source (355 nm, 600 ps pulse width, 3 kHz repetition rate) that is expanded and collimated before being directed onto a galvo-based fast-scanning axis and a motorized translation stage for the slow axis. The combination of the scan lens and scanning hardware provides a maximum field of view of 6 × 15 cm² and a scanning lateral pixel size of 7.5 µm. The fluorescence emission from the specimen is collected and routed into a multimode fiber for detection. The detection module contains avalanche photodetectors, dichroic beam splitters, and three spectral bands (390/40 nm, 470/28 nm, and 629/56 nm); see **Fig. 8**. Raw waveforms are digitized using a dual-channel high-speed acquisition module operating at 2.5 GS/s. Encoder outputs from the galvo and translation stage are captured by a data acquisition board to synchronize scanning and data readout, ensuring accurate spatial registration. The overall imaging speed is approximately 3,000 pixels/s and is limited by the repetition rate of the excitation source.

Human head and neck FFPE tissue sections from 19 patients were mounted on standard microscope slides and imaged without the use of coverslips. The samples were obtained from the Center for Genomic Pathology Laboratory at UC Davis. The tissue sections of head and neck tumor specimens were obtained under IRB #800853 (UC Davis). All the tissue sections were obtained after the de-identification of patient-related information and were prepared from existing (i.e., archived) specimens. Therefore, this work did not interfere with standard practices of care or sample collection procedures; the original tissue specimens were archived prior to this work and were not specifically collected for this project. Each field of view was acquired with the same scan parameters, spectral band configuration, and digitization settings.

Lifetime values were extracted from the digitized fluorescence waveforms using a constrained least-squares deconvolution method implemented with Laguerre expansion[47]. For each pixel, the raw waveform was background-corrected. The instrument response function, recorded for each spectral band, was used to deconvolve the measured signal. The algorithm yields intensity values (computed as the area under the waveform) and average lifetime estimates for all detection channels. The resulting lifetime maps across the three spectral channels ($LT_1$, $LT_2$, $LT_3$) together with their corresponding intensity images ($INT_1$, $INT_2$, $INT_3$) were used to form a six-channel representation of each tissue field of view.

All image data used for pixel super-resolution training and evaluation were derived from these high-resolution lifetime images, which served as the ground-truth reference for generating simulated low-resolution inputs. Lower resolution FLIM image data were generated by block-averaging the high-resolution FLIM images to simulate the reduced spatial sampling density associated with faster FLIM acquisition. The input for the $\text{FLIM}_{\text{PSR\_k}}$ network was generated by applying k × k non-overlapping block averaging to the high-resolution lifetime and autofluorescence channels. This corresponds to a lower resolution lateral pixel size of 7.5 × k µm.

**Dataset division and pre-processing**

To prepare the FLIM image data for training, we applied two pre-processing steps to the low and high-resolution image pairs. First, outliers were removed by clipping pixel values at the 99.5th percentile across all pixels within each channel of the low-resolution input. The high-resolution target images were clipped at the same per-channel values. Second, the images were min-max normalized. These pre-processing steps are critical for achieving stable training across different PSR models.

Whole slide images (WSIs) from 19 patients were included in the dataset. A strict patient-wise split was enforced to prevent subject leakage: WSIs from 16 patients were used for training, and WSIs from the remaining 3 held-out patients were reserved exclusively for testing. During both training and inference, each whole-slide FLIM image was partitioned into patches of 1920 × 1920 µm (i.e., 256 × 256 pixels for the high-resolution data). During training, these patches were provided to the network in pairs, with the average-pooling low-resolution image serving as the input and the corresponding high-resolution patch serving as the target (i.e., ground truth).

**cGAN model architecture and training**

Our cGAN pixel super-resolution framework consists of two distinct CNN-based architectures, namely a generator and a discriminator, which are jointly trained in an adversarial competition. The generator learns to perform super-resolution on the low-resolution inputs, while the discriminator learns to distinguish between ground-truth high-resolution images and super-resolved outputs from the generator. Our cGAN framework is inspired by the Least Squares Generative Adversarial Networks (LSGAN)[48], with a least squares objective that has a penalty based on the distance from the decision boundary between the generated and real image distributions, i.e.,

$$\mathcal{L}_{\text{discriminator}} = \left(D\big(G(I^{input})\big)\right)^2 + (D(I^{target}) - 1)^2 \quad (1)$$

$$\mathcal{L}_{\text{generator}} = \mathcal{L}_1^{\text{smooth}}\left(I^{target}, G(I^{input})\right) + \alpha \left(D\left(G(I^{input})\right) - 1\right)^2 \quad (2).$$

In our notation, $D(\cdot)$ and $G(\cdot)$ refer to the discriminator and generator networks, respectively. $\alpha$ is a hyperparameter that is empirically set to 0.1 for $k = 2$ and $k = 3$ PSR factors, and to 1 for $k \geq 4$.

As shown in Eq. (2), the generator loss function is augmented with a pixel-wise smooth L1 loss, also known as the Huber loss[49], which is defined as:

$$\mathcal{L}_1^{\text{smooth}} = \frac{1}{H \times W \times C} \sum_{i \in \{x,y,c\}} \ell_1 \left( [G(I^{input})]_i - [I^{target}]_i \right) \quad (3)$$

$$\ell_1(u) = \begin{cases} \frac{1}{2}u^2, & \text{if } |u| < 1 \\ |u| - 0.5, & \text{otherwise} \end{cases} \quad (4).$$

Here, $x$ and $y$ represent the lateral and vertical pixel index, and $c$ represents the image channel ($c \in \{\text{LT}_1, \text{INT}_1, \text{LT}_2, \text{INT}_2, \text{LT}_3, \text{INT}_3\}$). $H$, $W$, and $C$ represent the height, width and the number of image channels, respectively.

Our generator and discriminator architectures are shown in **Figure 9**. The generator employs a U-Net architecture preceded by a bilinear interpolation step. The U-Net encoder path consists of four levels of downsampling blocks. Each level contains 3 separate convolution blocks with batch normalization and ReLU activation. Residual connections are implemented at each level by zero-padding the channel dimension of the input and adding it to the output of the corresponding block. The first downsampling stage begins by increasing the number of input channels from 6 (3 lifetime and 3 corresponding autofluorescence intensity channels) to 64, which serves as the base channel count for the network, doubling at each subsequent level. These downsampling blocks are connected by average pooling layers with a stride of two, which reduce the output of the previous block by a factor of 2 in both lateral dimensions. Thus, at each encoder level, the number of channels doubles while the spatial dimensions decrease. The decoder path follows a reverse structure, comprising 4 levels of upsampling blocks. Like the encoder, each block contains three convolutional layers, followed by batch normalization and an activation function, and receives the output of the previous level concatenated with the corresponding encoder output via skip connections. Consecutive upsampling levels are connected by bilinear interpolation, which increases each lateral dimension by a factor of 2. At the final decoder level, a convolution layer produces the generated image with 6 output channels.

The discriminator network consists of an initial convolution layer followed by 5 discriminator blocks. Each block consists of two convolutional layers with batch normalization and ReLU activation. The first convolution preserves both channel count and spatial dimensions, while the second doubles the number of channels and halves the lateral dimensions through strided convolution. Similar to the generator, the base feature channel is set to 64 and doubles after each block. Finally, two fully connected layers with a sigmoid activation output the probability of the input being real or fake.

**Brownian-bridge diffusion model (BBDM)**

We also trained a conditional Brownian bridge stochastic diffusion process[42] that translates from the low-resolution lifetime and autofluorescence domain $I^{input} \in \mathbb{R}^{\frac{H}{k} \times \frac{W}{k} \times 6}$ to the corresponding super-resolution domain $x_0 = I^{target} \in \mathbb{R}^{H \times W \times 6}$, where $k$ is the super-resolution factor. The low-resolution images are first upsampled to match the target spatial dimensions:

$$x_T = y = f_k(I^{input}) \tag{5}$$

where $f_k$ denotes bilinear interpolation with upsampling factor $k$, and T represents the total number of diffusion steps (set to 1000 for both training and testing). During the forward process, Gaussian noise is progressively added to the high-resolution image $x_0$. At each timestep $t$, the noisy image $x_t$ follows a Gaussian distribution with mean $(1 - m_t)x_0 + m_t y$, and variance $\delta_t = 2s(m_t - m_t^2)$, where $s$ is a hyperparameter set to 1 and $m_t = \frac{t}{T}$. In other words, the higher-resolution ground truth $x_0$ is progressively corrupted into an interpolated low-resolution image $x_T$ through a forward Brownian bridge process, which is fixed at both ends. During the reverse process, a denoiser attention U-Net predicts the Gaussian noise added to $x_0$ at timestep $t$ in the forward pass. The forward equations and reverse denoising procedure are summarized in the **Supplementary Note**. The original BBDM framework applies diffusion in the latent space without conditioning on the input[42]. In this work, we directly apply the process to the space of lifetime and autofluorescence intensity channels while passing low-resolution channels as conditions to the denoising U-Net. The denoiser U-Net architecture is detailed in the **Supplementary Note**.

**Performance evaluation metrics**

To quantitatively evaluate the performance of the proposed FLIM$_{PSR\_k}$ framework, we employed complementary metrics that capture pixel-level accuracy, structural similarity, and perceptual similarity. All quantitative metrics were computed separately for each FLIM modality (intensity and lifetime) and for each imaging channel. The final reported values were obtained by averaging across channels. In addition to these quantitative evaluations, we performed a qualitative analysis focusing on image features of biological relevance, such as tissue boundaries and small-scale structural variations. Representative examples of these qualitative comparisons are shown in **Figures 2** and **3**, where the cross-sectional profiles highlight the similarity between the reconstructed images and their high-resolution references.

The MSE was used to quantify the average pixel-wise difference between the super-resolved FLIM output and its corresponding high-resolution ground truth. Given a predicted FLIM$_{PSR\_k}$ image $\hat{I}$ and the high-resolution FLIM image $I$, the MSE is defined as

$$\text{MSE} = \frac{1}{N} \sum_{i=1}^{N} (\hat{I}_i - I_i)^2 \tag{6}$$

where $N$ denotes the total number of pixels in the image. The PSNR was further derived from the MSE as

$$\text{PSNR} = 10\log_{10}\left(\frac{L^2}{\text{MSE}}\right) \tag{7}$$

where $L$ denotes the maximum possible pixel value (for normalized images, $L = 1$) and PSNR expresses super-resolved fidelity on a logarithmic scale, where larger values correspond to smaller errors.

To evaluate structural consistency, we employed the SSIM[41], which assesses luminance, contrast, and structural correlation between the super-resolved output and the corresponding high-resolution FLIM ground truth. Given a predicted FLIM$_{PSR\_k}$ image $\hat{I}$ and the high-resolution FLIM image $I$, the SSIM is defined as

$$\text{SSIM}(\hat{I}, I) = \frac{(2\mu_{\hat{I}}\mu_I + C_1)(2\sigma_{\hat{I}I} + C_2)}{(\mu_{\hat{I}}^2 + \mu_I^2 + C_1)(\sigma_{\hat{I}}^2 + \sigma_I^2 + C_2)}, \tag{8}$$

where $\mu_{\hat{I}}$ and $\mu_I$ denote the image means, $\sigma_{\hat{I}}^2$ and $\sigma_I^2$ denote the variances, $\sigma_{\hat{I}I}$ denotes the covariance between the two images, and $C_1$ and $C_2$ are small constants to stabilize the division.

In addition to pixel-level and structural measures, we employed the LPIPS[40] metric to quantify perceptual fidelity. LPIPS computes the distance between deep feature representations extracted from a pretrained visual geometry group (VGG) network, providing a perceptually aligned measure of similarity between the super-resolved and ground-truth images. Lower LPIPS scores correspond to images that are perceptually closer to the reference.

To verify true resolution enhancement rather than simple edge sharpening, we also performed a spatial frequency spectrum analysis on both the lifetime and intensity modalities (**Figures 2L** and **3L**). For each super-resolved FLIM image, a two-dimensional Fourier Transform was applied directly to the normalized output to obtain its frequency-domain representation. The radially averaged power spectral density was then computed to quantify the recovery of high-frequency components across the low-resolution, super-resolved, and ground-truth images. This analysis provides a measure of the effective frequency-domain resolution gain achieved by the model.

**Statistical analysis**

A paired two-sided t-test was used to evaluate whether diffusion-based models produced statistically significant differences in reconstruction quality relative to the cGAN model. This analysis was performed for three image-quality metrics (SSIM, PSNR, and LPIPS), computed independently for each of the six FLIM channels and for super-resolution factors of k = 5, 6 and 7. For each condition, metric values from the diffusion and cGAN outputs were paired across 128 fields of view from the test dataset.

The null hypothesis stated that both models yielded outputs with identical mean performance. A significance threshold of α = 0.05 was used, with equal allocation for superiority and inferiority due to the two-sided nature of the test. Results were interpreted according to the sign of the t-statistic and the associated p-value: statistically significant improvement of cGAN performance over the performance of the diffusion model was assigned when $p \leq 0.05$ and $t > 0$ for SSIM and PSNR (or $t < 0$ for LPIPS, for which lower values indicate better performance); statistically significant degradation of cGAN relative to diffusion was assigned when $p \leq 0.05$ and the t-statistic had the opposite sign; and results with $p > 0.05$ were considered not statistically significant. These outcomes are shown in **Fig. 6** and **Fig. S7**, which demonstrate that diffusion models consistently underperformed relative to the cGAN at $k = 5$, whereas at higher super-resolution factors diffusion models exhibited statistically significant advantages for some channels and metrics, particularly for SSIM at $k = 7$.

**Other implementation details**

All FLIM deconvolution, lifetime reconstruction, and pre-processing steps were carried out in MATLAB (R2023b). Network training and evaluation were run on a workstation featuring an AMD Ryzen 9 5900X processor, 128 GB RAM, and an NVIDIA GeForce RTX 3090Ti GPU. The model architecture, training, and inference algorithms were implemented in Python 3.9.19 with PyTorch 2.2.1.

**Supplementary Information:** This file contains the Supplementary Note and Supplementary **Figures S1-S6**.

**Figures**

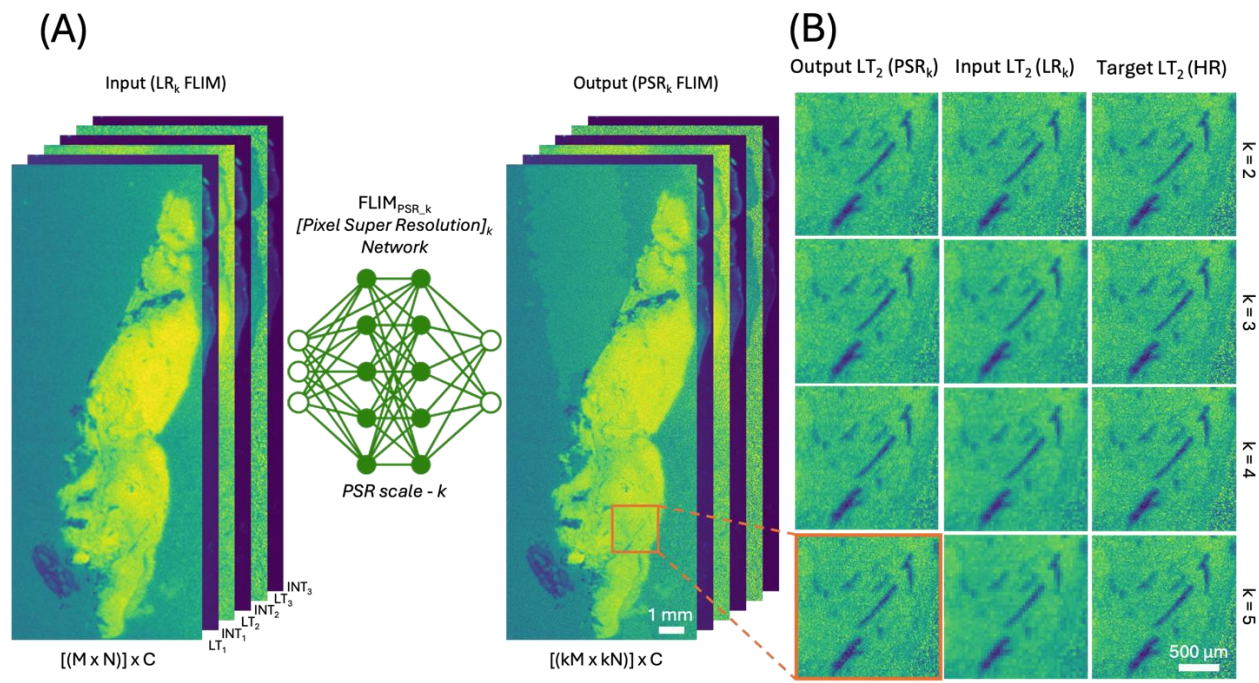

**Figure 1:** FLIM Pixel Super-Resolution Framework Overview. **(A)** Schematic of the FLIM$_{PSR\_k}$ framework. A low-resolution FLIM input with spatial size (M×N) and C channels (e.g., C = 6) is mapped by the FLIM$_{PSR\_k}$ network to a super-resolved output with spatial size (kM × kN) and the same number of channels (C), where k is the PSR factor (e.g., k = [2, 3, 4, 5]). The high-resolution FLIM data serve as the target during training. **(B)** Example field of view of a head and neck tissue section showing lifetime channel LT$_2$.

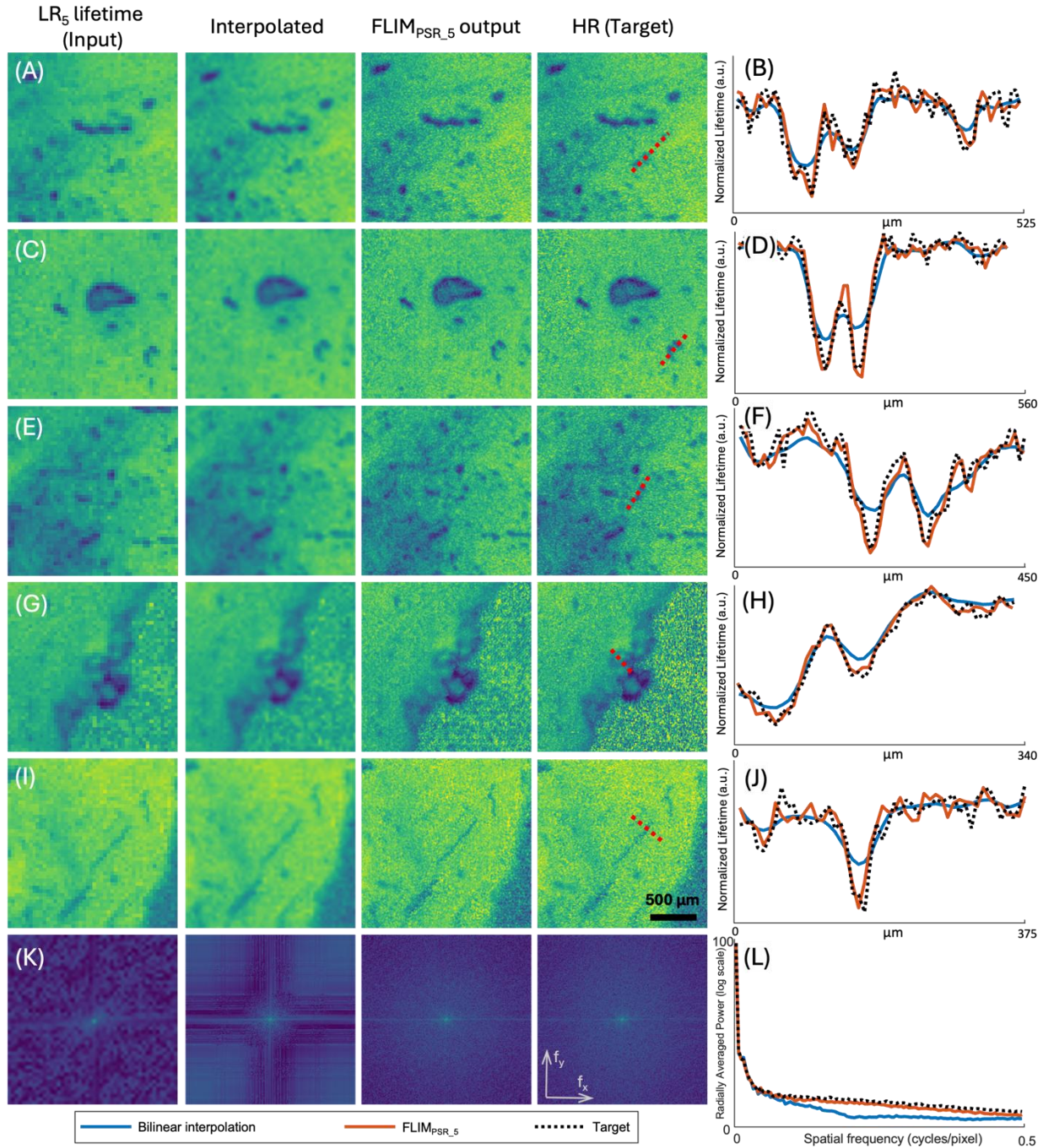

**Figure 2:** Structural Fidelity and Resolution Validation of FLIM$_{PSR\_5}$ Lifetime Images. **(A, C, E, G, I)** Representative lifetime channel examples at k = 5 PSR. Low-resolution inputs (LR$_5$), interpolated images, FLIM$_{PSR\_5}$ outputs, and high-resolution target images. All images are shown with the same lifetime colormap and scaling. The shown fields of view correspond to channels LT$_2$, LT$_2$, LT$_1$, LT$_3$, and LT$_1$, respectively. **(B, D, F, H, J)** Line profiles extracted along the paths indicated in the HR target regions. The FLIM$_{PSR\_5}$ output image profiles (solid red lines) closely follow the HR target traces (dashed black lines), recovering sharp lifetime transitions and local

gradients, while the interpolated profiles (solid blue lines) remain smoothed and under-represent local variation. **(K)** Spatial frequency domain of all images shown in (I), and **(L)** the radially averaged spatial frequency spectra for the low resolution, $FLIM_{PSR\_5}$, and HR lifetime. The $FLIM_{PSR\_5}$ outputs recover mid- and high-frequency content and closely match the HR spectrum. Also see **Figure S1** for additional examples.

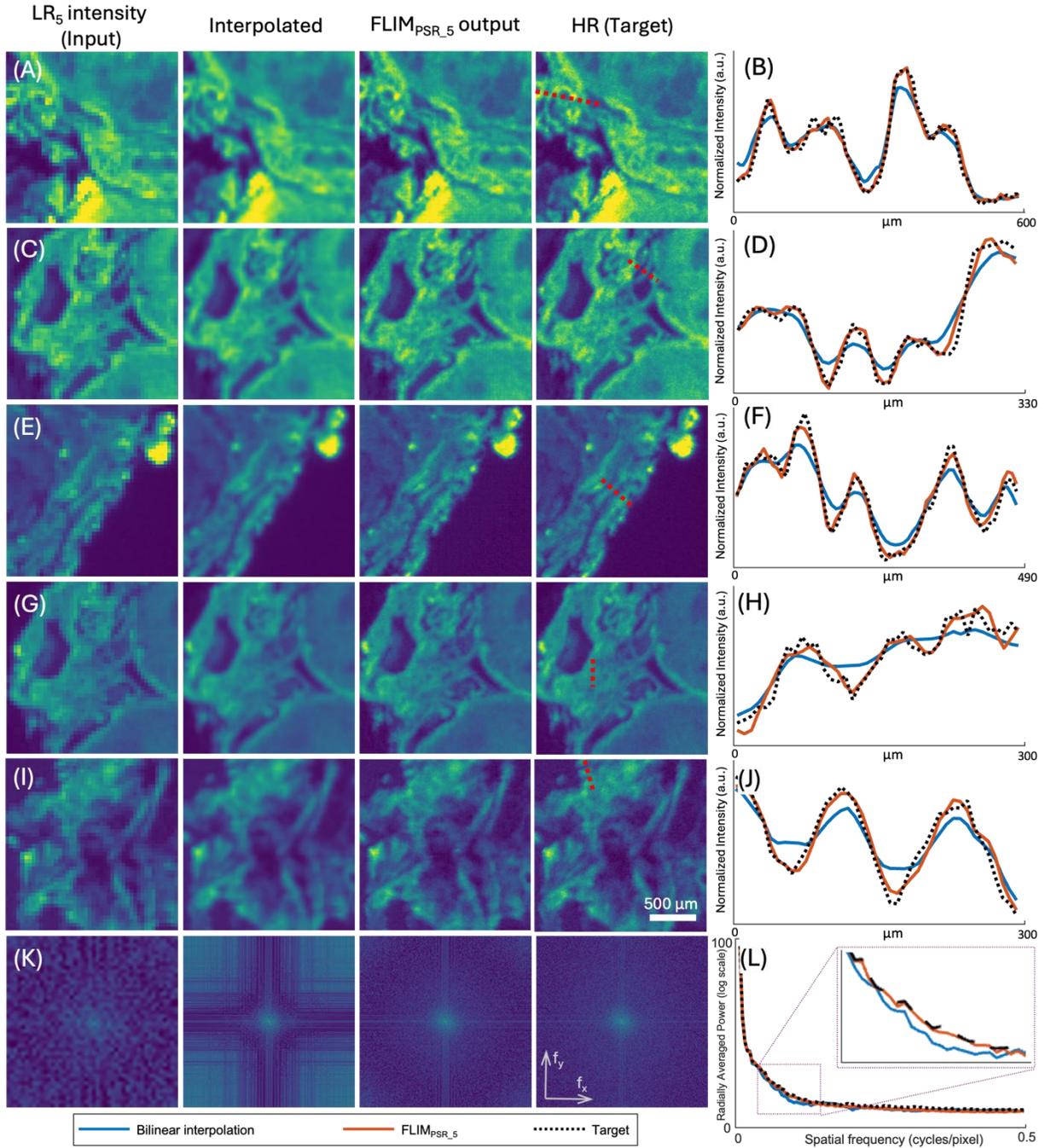

**Figure 3:** Structural Fidelity and Resolution Validation of FLIM$_{PSR\_5}$ Intensity Images. **(A, C, E, G, I)** Representative intensity channel examples at k = 5 PSR. Low-resolution inputs (LR$_5$), interpolated images, FLIM$_{PSR\_5}$ outputs, and high-resolution target images. All images are shown with the same intensity colormap and scaling. The shown fields of view correspond to channels INT$_1$, INT$_1$, INT$_2$, INT$_2$, and INT$_1$, respectively. **(B, D, F, H, J)** Line profiles extracted along the paths indicated in the HR target regions. The FLIM$_{PSR\_5}$ output image profiles (solid red lines) closely follow the HR target traces (dashed black lines), recovering sharp intensity transitions and local gradients, whereas the interpolated profiles (solid blue lines) remain smoothed and underrepresent local variation. **(K)** Spatial frequency domain of all images shown in (I) and **(L)** the radially averaged spatial frequency spectra for the low resolution, FLIM$_{PSR\_5}$, and HR intensity. The FLIM$_{PSR\_5}$ outputs recover mid-frequency content and closely match the HR spectrum. Also see **Figure S2** for additional examples.

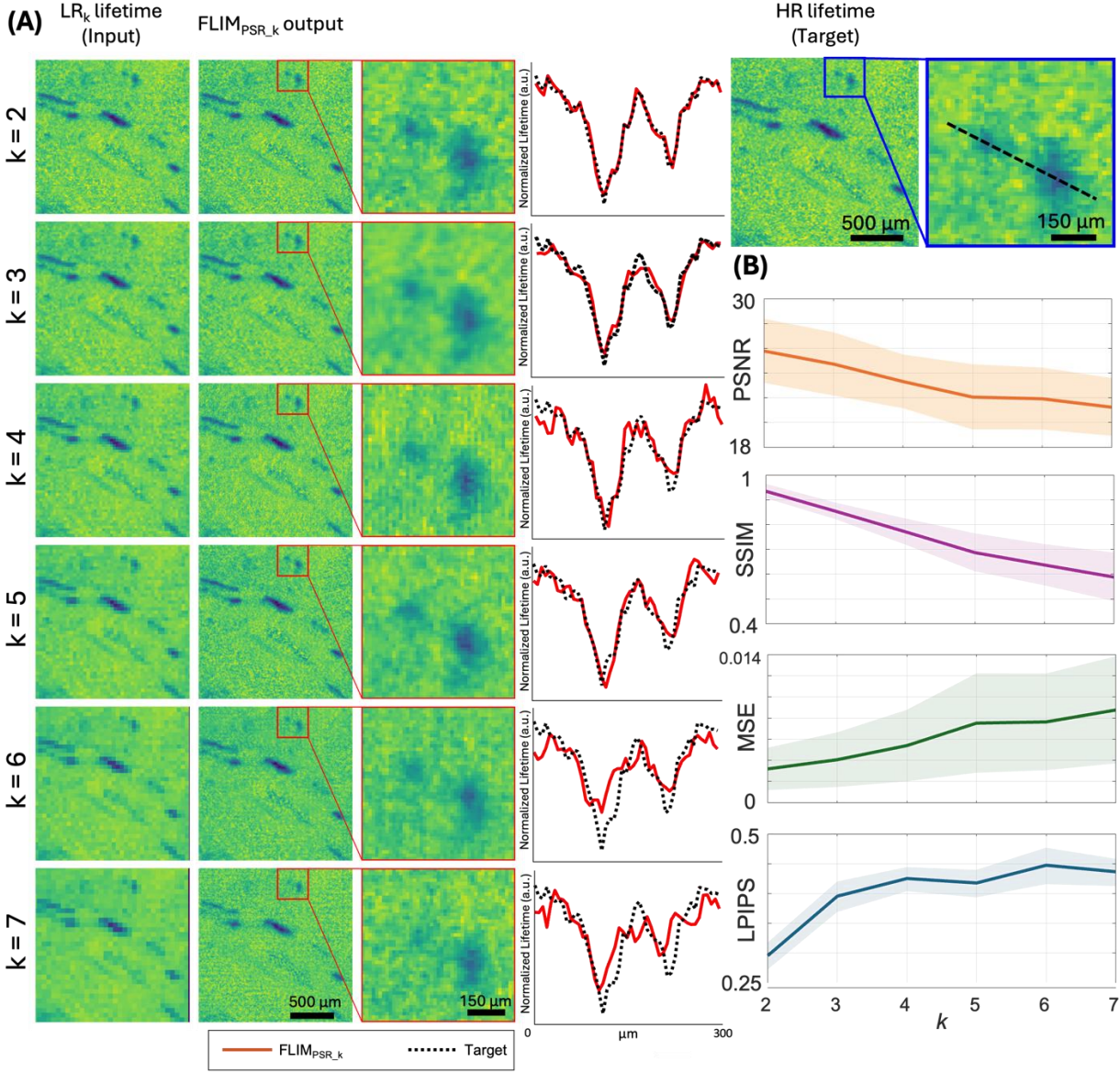

**Figure 4:** Quantitative Performance of Lifetime FLIM$_{PSR\_k}$ Across PSR Factors. **(A)** Lifetime reconstructions for a representative field of view at PSR factors of k = 2,3,4,5,6 and 7. For each k, the panel shows the low-resolution input (LR) and the FLIM$_{PSR\_k}$ output, together with the HR lifetime target. Up to k=5, the PSR outputs retain clear tissue boundaries and local lifetime variations that match the HR reference FLIM image. Zoomed-in regions from the same field of view for all PSR factors are shown in the third column. Line profiles through the zoomed feature are shown for each PSR factor. The field of view shown corresponds to LT$_2$. **(B)** PSNR, SSIM, MSE, and LPIPS metrics of the full test dataset as a function of the PSR factor k for all lifetime channels (shaded regions indicate the standard deviations). Also see **Figure S3** for additional examples.

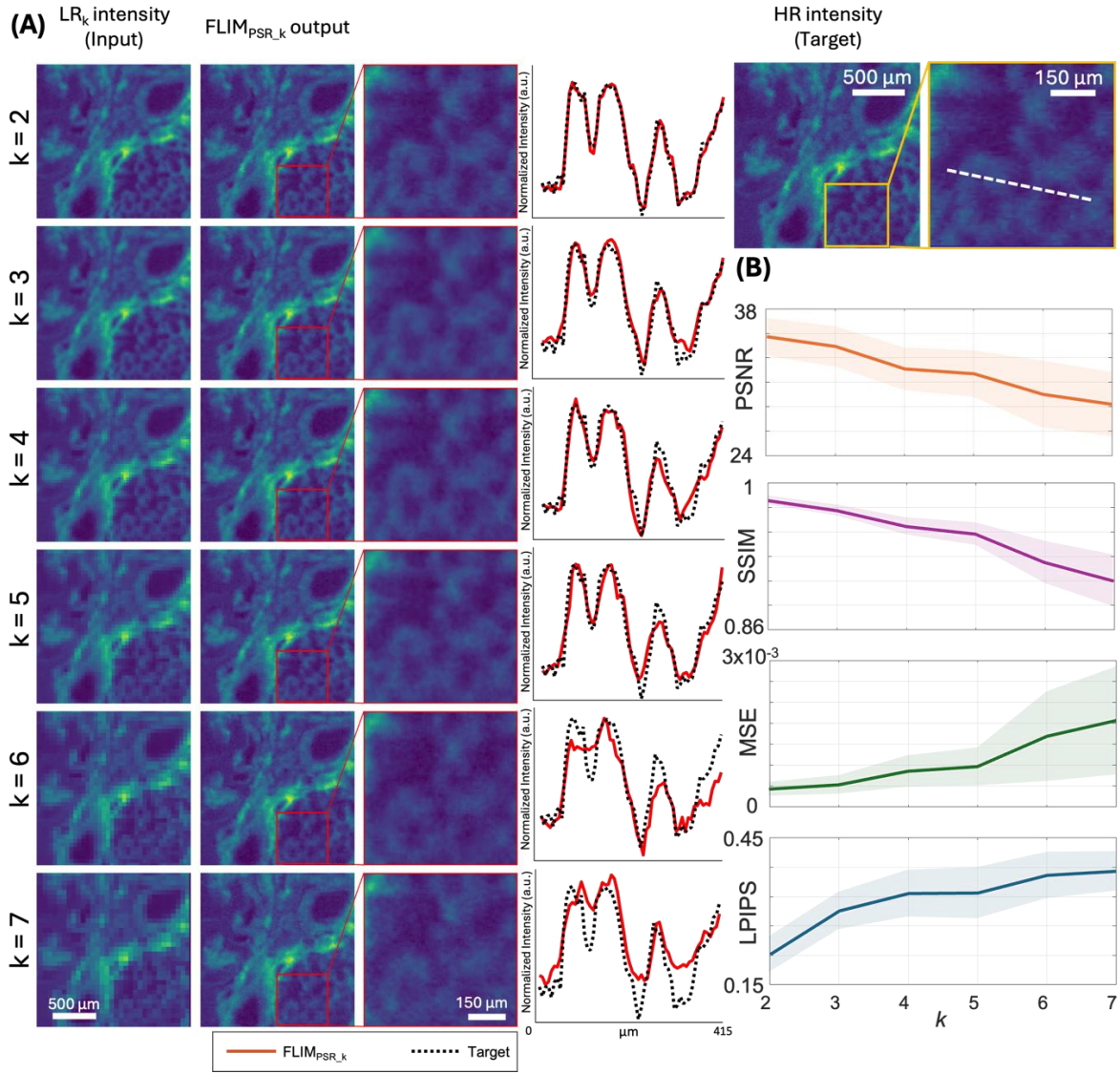

**Figure 5:** Quantitative Performance of Intensity FLIM$_{PSR\_k}$ Across PSR Factors. **(A)** Intensity reconstructions for a representative field of view at PSR factors of k = 2,3,4,5,6 and 7. For each k, the panel shows the low-resolution input (LR) and the FLIM$_{PSR\_k}$ output, together with the HR intensity target. Up to k = 5, the PSR outputs retain clear tissue boundaries and local intensity variations that match the HR reference. Zoomed-in regions from the same field of view for all PSR factors are shown in the third column. Line profiles through the zoomed feature are shown for each PSR factor. The field of view shown corresponds to INT1. **(B)** PSNR, SSIM, MSE, and LPIPS metrics of the full test dataset as a function of the PSR factor k for all intensity channels (shaded regions indicate the standard deviations). Also see **Figure S4** for additional examples.

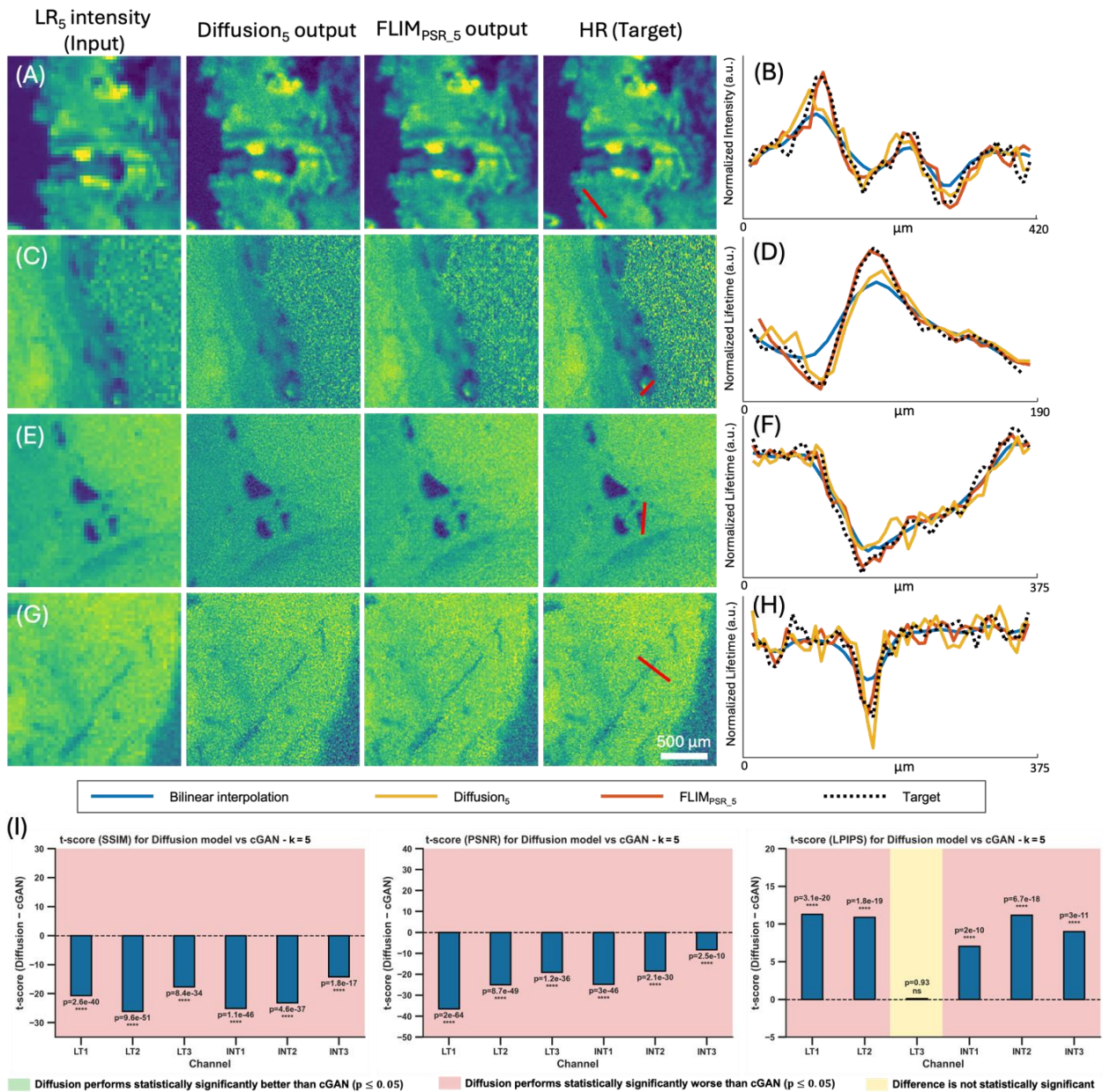

**Figure 6:** Comparison of cGAN and Diffusion-Based FLIM PSR Image Reconstructions at k = 5. **(A, C, E, G)** Representative intensity and lifetime channel examples at k = 5. Low-resolution input (LR$_5$), Diffusion$_5$ output, FLIM$_{PSR\_5}$ output, and HR target. All images of the same FOV are shown with the same colormap and scaling. The shown fields of view correspond to channels INT$_1$, LT$_2$, LT$_2$, and LT$_1$, respectively. **(B, D, F, H)** Line profiles extracted along the paths indicated in the HR target regions. **(I)** SSIM, PSNR, and LPIPS metric comparisons between the diffusion and cGAN test datasets for each lifetime and intensity channel. For all FLIM channels, the cGAN-based model outperforms the diffusion-based model. All results are statistically significant, except for the LPIPS metric in channel LT$_3$.

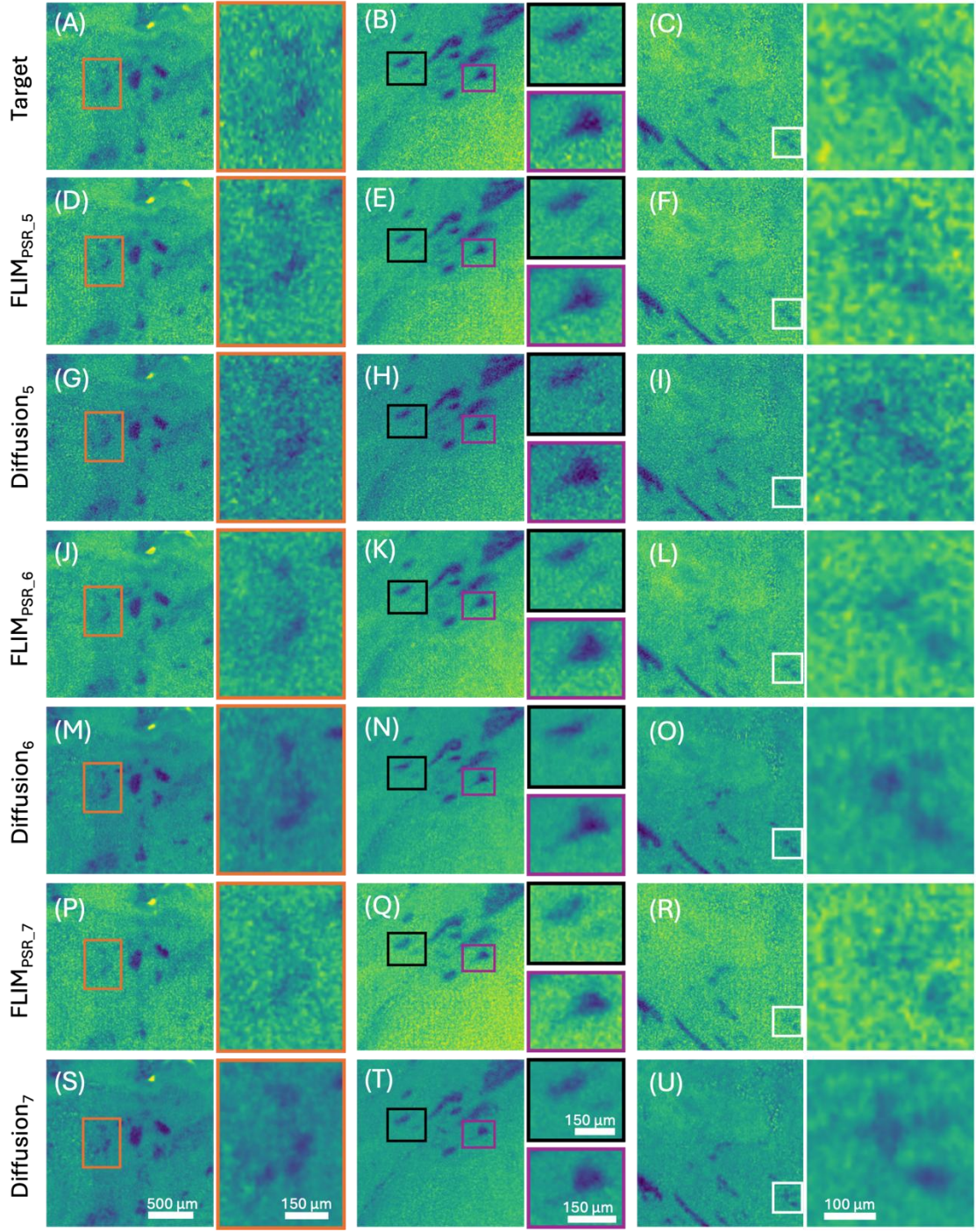

**Figure 7:** Comparative Analysis: Diffusion vs. cGAN (higher k regime). Target **(A - C)**, FLIM$_{PSR\_5}$ **(D - F)**, Diffusion$_5$ **(G - I)**, FLIM$_{PSR\_6}$ **(J - L)**, Diffusion$_6$ **(M - O)**, FLIM$_{PSR\_7}$ **(P - R)**, and Diffusion$_7$ **(S - U)** of lifetime fields of view. Zoomed-in visualizations of selected features of interest are shown to the right of each field of view. At PSR factors above 5, the cGAN-based reconstructions show diminished fine-feature recovery, whereas the diffusion model-based reconstructions tend toward over-sharpening.

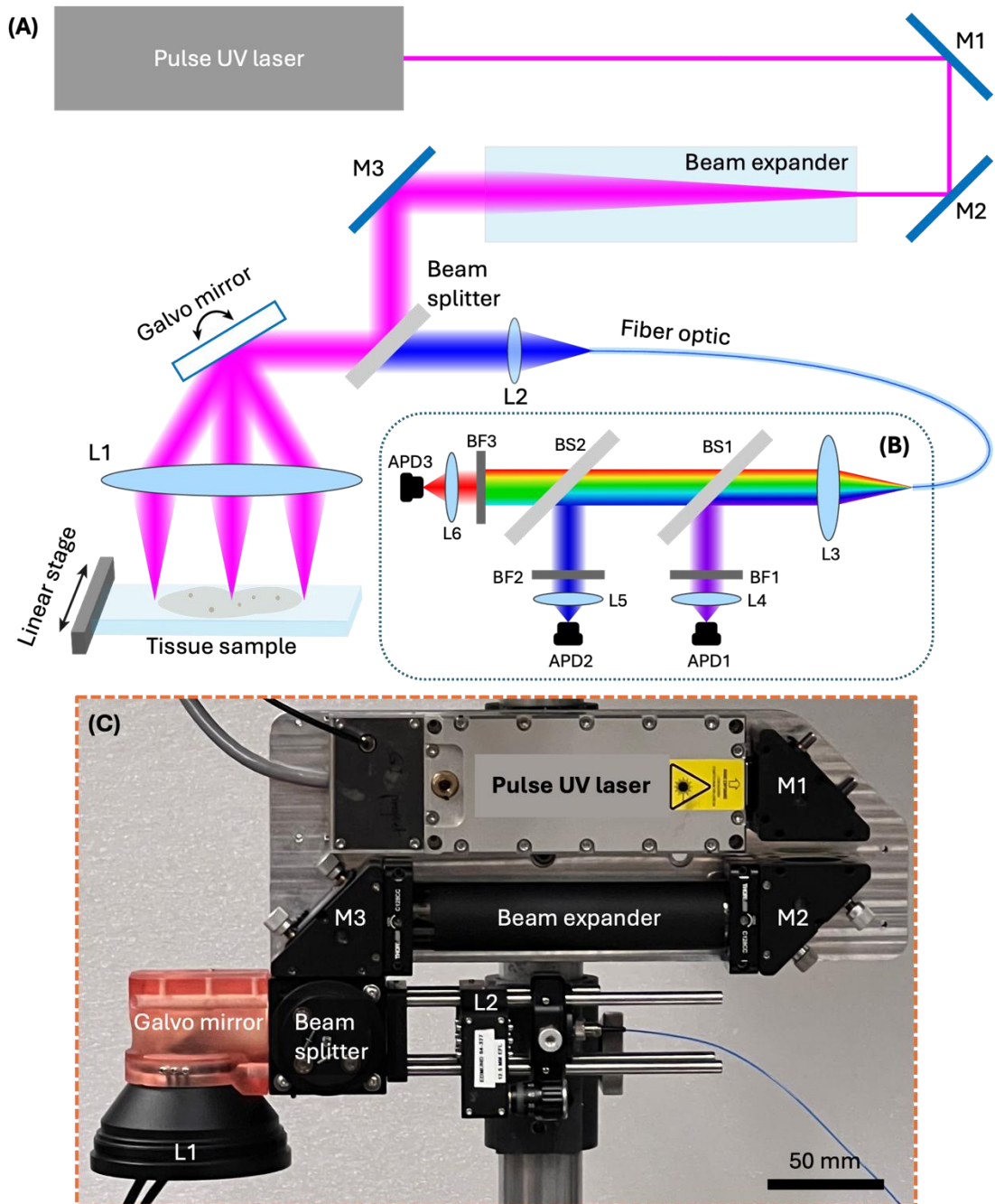

**Figure 8:** LSPS-FLIM Optical Path and Design. **(A)** Schematic of the scanning FLIM system. The system utilizes a 355 nm pulsed UV laser operating at 3 kHz, producing 600 ps pulses with an energy of 0.2 µJ. After emission, the beam is routed through a pair of dielectric mirrors (M1 and M2) and subsequently expanded and collimated by a 10× beam expander. A third dielectric mirror (M3) and an ultra-flat dichroic beam splitter redirect the collimated beam toward the scanning assembly. Fast-axis scanning is driven by a galvo mirror, and an F-theta lens (L1, 60 × 60 mm² field, 100 mm focal length, 12 mm entrance diameter) focuses the beam while maintaining uniform resolution across a 60 mm scan range. Slow-axis motion is provided by a motorized brushless linear translation stage, enabling up to 150 mm of travel. Fluorescence generated at the sample is returned along the same optical path to the galvo mirror and then coupled, via a convex lens (L2), into a multimode collection fiber (365 µm core, 0.22 NA). **(B)** Detection is carried out by a FLIM module incorporating avalanche photodetectors, dichroic splitters, and bandpass filters[50]. The system captures three spectral channels centered at 390/40, 470/28, and 629/56 nm. **(C)** Picture of the illumination section of the FLIM scanning system.

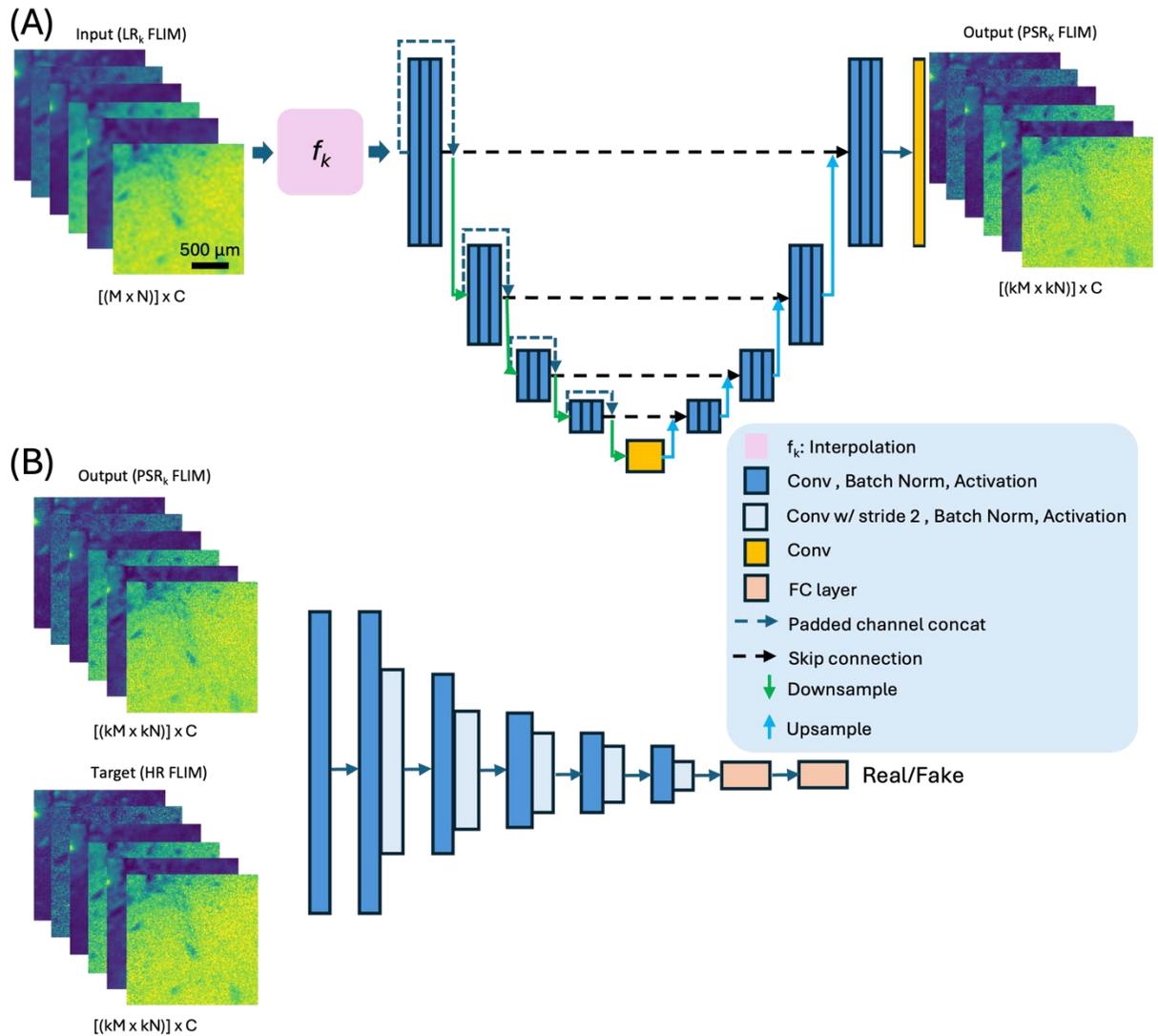

**Figure 9:** Model Architecture. A cGAN framework consisting of a generator model and a discriminator model was used to train the FLIM$_{PSR\_k}$ networks. **(A)** The generator network follows a U-Net architecture with 4 levels of downsampling blocks in the encoder and 4 levels of upsampling blocks in the decoder. Each level comprises three separate convolutional blocks, each with batch normalization and ReLU activation. Residual connections are implemented at each level by zero-padding the channel dimension of the input and adding it to the output of the corresponding block. The model receives a low-resolution FLIM input of size $[(M \times N)] \times C$, which is interpolated to $[(kM \times kN)] \times C$ at the input of the generator. The generator outputs a PSR$_k$ image of size $[(kM \times kN)] \times C$. **(B)** The discriminator network consists of an initial convolution layer followed by 5 discriminator blocks. Each block consists of two convolutional layers with batch normalization and ReLU activation. The first convolution preserves both channel count and spatial dimensions, while the second one doubles the number of channels and halves the lateral dimensions through strided convolution.

# REFERENCES

1   Lakowicz, J. R. in *Principles of Fluorescence Spectroscopy* (ed Joseph R. Lakowicz) 741-755 (Springer US, 2006).

2   BECKER, W. Fluorescence lifetime imaging – techniques and applications. *Journal of Microscopy* **247**, 119-136 (2012). https://doi.org:https://doi.org/10.1111/j.1365-2818.2012.03618.x

3   Suhling, K. *et al.* Fluorescence lifetime imaging (FLIM): Basic concepts and some recent developments. *Medical Photonics* **27**, 3-40 (2015). https://doi.org:https://doi.org/10.1016/j.medpho.2014.12.001

4   Liu, X., Bec, J., Zhou, X., Garcia, A. A. & Marcu, L. Multispectral laser-scanning pulse-sampling fluorescence lifetime system for large-scale tissue imaging. *Opt Lett* **50**, 900-903 (2025). https://doi.org:10.1364/ol.547582

5   Redford, G. I. & Clegg, R. M. Polar plot representation for frequency-domain analysis of fluorescence lifetimes. *J Fluoresc* **15**, 805-815 (2005). https://doi.org:10.1007/s10895-005-2990-8

6   Stringari, C. *et al.* Phasor approach to fluorescence lifetime microscopy distinguishes different metabolic states of germ cells in a live tissue. *Proc Natl Acad Sci U S A* **108**, 13582-13587 (2011). https://doi.org:10.1073/pnas.1108161108

7   Suhling, K. *et al.* Wide-field time-correlated single photon counting-based fluorescence lifetime imaging microscopy. *Nucl Instrum Methods Phys Res A* **942**, 162365 (2019). https://doi.org:10.1016/j.nima.2019.162365

8   Liu, X. *et al.* Fast fluorescence lifetime imaging techniques: A review on challenge and development. *Journal of Innovative Optical Health Sciences* **12**, 1930003 (2019). https://doi.org:10.1142/s1793545819300039

9   Kanno, H. *et al.* High-speed fluorescence lifetime imaging microscopy: techniques, applications, and prospects. *Biophotonics Discovery* **2**, 030901 (2025).

10  Fernandes, S. *et al.* Fibre-based fluorescence-lifetime imaging microscopy: a real-time biopsy guidance tool for suspected lung cancer. *Translational Lung Cancer Research* **13**, 355-361 (2024).

11  Alfonso-Garcia, A. *et al.* In vivo characterization of the human glioblastoma infiltrative edge with label-free intraoperative fluorescence lifetime imaging. *Biomedical Optics Express* **14**, 2196-2208 (2023). https://doi.org:10.1364/BOE.48130426